\documentclass{bmvc2k}
\usepackage{tikz}
\usepackage{svg}
\usepackage{graphicx}
\usepackage{amsmath}
\usepackage{amssymb}
\usepackage{booktabs}
\usepackage{multirow}
\usepackage{hhline}
\usepackage[parfill]{parskip}

\usepackage{standalone}
\usepackage{standalone}
\usepackage[activate={true,nocompatibility},final,tracking=true,kerning=true,spacing=true,factor=1100,stretch=10,shrink=9]{microtype}

\title{Two-Stream Transformer Architecture for Long Form Video Understanding}

\addauthor{Edward Fish}{edward.fish@surrey.ac.uk}{1}
\addauthor{Jon Weinbren}{j.weinbren@surrey.ac.uk}{1}
\addauthor{Andrew Gilbert}{a.gilbert@surrey.ac.uk}{1}

\addinstitution{
 The University of Surrey
}

\begin{document}
\maketitle

\begin{abstract}
Pure vision transformer architectures are highly effective for short video classification and action recognition tasks. However, due to the quadratic complexity of self attention and lack of inductive bias, transformers are resource intensive and suffer from data inefficiencies. Long form video understanding tasks amplify data and memory efficiency problems in transformers making current approaches unfeasible to implement on data or memory restricted domains. This paper introduces an efficient Spatio-Temporal Attention Network (STAN) which uses a two-stream transformer architecture to model dependencies between static \textit{image} features and temporal \textit{contextual} features. Our proposed approach can classify videos up to two minutes in length on a single GPU, is data efficient, and achieves SOTA performance on several long video understanding tasks.

\end{abstract}

\section{Introduction}
\label{sec:intro}
Long form video understanding (LVU) is a sub-domain of video recognition concerned with understanding contextual information across contiguous shots which can contain multiple locations, scenes, interactions, and actions. While a blurb can give a snapshot of a book, stories are enriched by the development of characters and their interactions with objects, people, and locations. The same applies in video recognition where understanding the context of individual moments in relation to a whole video can provide valuable information for tasks such as classification, speaker recognition, character understanding, and video retrieval. However, current video recognition methods and datasets have tended to focus on short videos \cite{hollywood,vlog,3dconv,cascante2019moviescope,movienet,zhou2010movie,multimodalfusion-video-attention,x3d,pseudo3d}, effectively aggregating convolutional image features via late fusion and inflation \cite{x3d,largescale,3dconv,pseudo3d,channel-seperated}, or extending attention based image classification methods to the tasks of action recognition, object tracking, and segmentation \cite{videobert,vilbert,axial-attention,Anticipate-vid-transformer,TimesFormer,attention-nas}. 

 \begin{figure}[htbp]
 \label{fig:overview}
  \centering
  \includegraphics[width=0.9\textwidth]{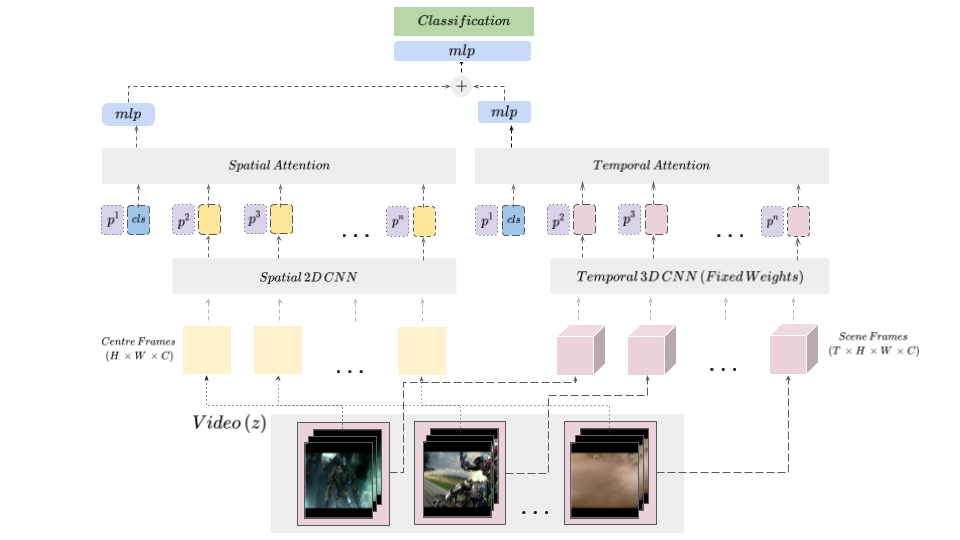}
  
  \caption{An overview of our approach, STAN. We encode video scenes into two feature representations using a two-stream spatio-temporal convolutional network. A transformer encoder is then used to model temporal dependencies between the tokens via an additional classification token randomly initialised. The proposed method allows us to model long term dependencies between individual frames of long videos which feature multiple actions and environments.}
 \end{figure}
 
Intuitively, an image can provide a good summary of a moment in time, and so it is logical that these methods perform well at classifying short videos from only a few frames \cite{beyond_snippets}; however, many video classification tasks require, or can be improved with, long-term temporal reasoning \cite{wu2021towards}. The temporal fusion of features via recurrent networks \cite{rnn,lstm} provides one solution by aggregating frames over longer sequences of frames, but these architectures suffer from computational inefficiencies due to their recursive design while 3D convolutional approaches also do not scale effectively, primarily due to their high computational footprint and the linear growth of the receptive field in the temporal dimension.  

More recently, transformers \cite{bert,16x16} have been adapted to the video domain \cite{vivit,videobert,Anticipate-vid-transformer}, achieving SOTA results on multiple short video tasks. But due to the lack of inductive bias in transformer architectures, it has been observed that video transformers are not as data efficient as their CNN counterparts \cite{DEITT}, requiring more data and thus more time to train. Furthermore, it has been shown that transformer image networks trained on natural videos work just as effectively when inputs are shuffled \cite{multiscale}, demonstrating weak temporal bias. This issue is compounded when addressing LVU tasks, as we encounter larger datasets with fewer opportunities for parameter updates during training caused by a small target to data ratio. 

The following question then arises.\textbf{How can we leverage inductive bias from image CNN's to make video transformer networks more data and memory efficient for long form video understanding tasks?} 

Our solution is a two-stream Spatio-Temporal Attention Network (STAN) with which we gain data and computational efficiency by introducing inductive bias via convolution. Existing methods for image classification with transformers such as \cite{16x16} split images into $16 \times 16$ pixel regions encoded with a positional embedding to introduce permutation and translation invariance. This method has been extended to video by expanding these regions temporally to create 3D tokens with 2D positional embeddings \cite{vivit}. Inspired by work in two-stream slow-fast convolutional video networks for short video classification \cite{slowfast}, our approach replaces this tokenisation method with both \textit{image} spatial and \textit{context} temporal scene features extracted from pre-trained convolutional neural networks. We then use a two-stream transformer architecture to model temporal dependencies between the scene features for classifying long videos of up to two minutes in length. Our method is data and memory efficient and achieves SOTA results on a range of long form video understanding tasks. 

\section{Related Work}
\label{sec:literature-review}

Early video classification works focused on non-temporal aggregation, which included clustering \cite{vlad,fisher,learnablepooling} or fusing \cite{rethinking-genre} spatial features obtained from convolutional neural networks. Since a short video will share a similar distribution of pixels over concurrent frames, these networks perform well for video classification and object detection tasks. Naturally, fusing these output features using RNN's \cite{rnn,sequential,faster-rcnn,hybrid-network} such as an LSTM \cite{lstm,beyond_snippets} improves performance by introducing temporal information. However, as discussed in \cite{16x16}, these model architectures align with steps in computational time, and are inefficient at longer sequence lengths as memory constraints limit batching across sequences. 

Later works explored extending convolution to video, inflating image CNN's via a temporal channel \cite{x3d,largescale,3dconv,pseudo3d,carreira2017quo_kinetics} and using two stream convolutional networks to aggregate spatial and temporal information introduced via optical flow \cite{two-stream}, or from various sampling and aggregation intervals \cite{r21d,actorcontextactor,spati-temp-speed-accuracy}. 3D Convolutional Neural Networks (CNN's) are highly effective at video classification, object detection, and action recognition tasks but are computationally intensive to train. For example, in \cite{threedCNN} the authors process just 16 frames at the cost of 40 GFLOPS per single pass, making the approach infeasible for sequences longer than a few seconds. To address computational overhead in training convolutional video networks, in \cite{slowfast} the authors present an efficient and high performance CNN for video classification, which utilises a fast and slow temporal sampling stream. The slow stream can process higher resolution images and extract key spatial information over a few key frames, while the fast stream maps low resolution frames to infer temporal information. This method is highly effective at video and action classification on short 10 second clips at a low computational cost.

Transformers \cite{16x16, bert,attention_all_you_need,gpt3} use self-attention to model dependencies between inputs and have recently been shown to work effectively for video classification tasks when implemented with temporal positional information \cite{videobert,vilbert,axial-attention,Anticipate-vid-transformer,TimesFormer,attention-nas}. Unlike convolutional methods, they lack inductive bias and, as such, take longer to model dependencies between neighbouring pixel regions which extends training time and harms data efficiency \cite{DEITT} thus making them innapropriate for LVU tasks. Introducing inductive bias via convolution, shifting windows, and gaussian bias to transformers has shown to be effective in the image domain \cite{token-to-token,cmt-convolutions-meet-transformers,residualattention,localness-for-self-attention,tsm}, and hybrid networks for video have also been proposed in \cite{video-transformer-network,convit,video-action-transformer}, but only on short segments of video. Finally, LVU networks have been explored in \cite{wu2021towards, zhao2019long} where the authors leverage short-term detection and tracking to form instance representations or use LSTM's for the task of long form video-question answering respectively.


\section{Method}
\label{method}                              
\begin{figure}[htpb]
 \centering
  \includegraphics[width=0.8\textwidth]{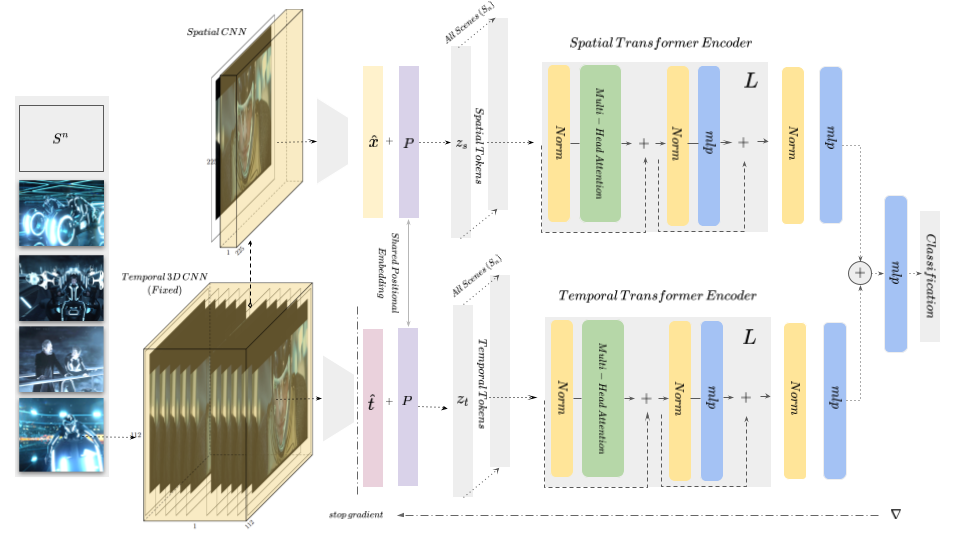} 
  \caption{For each scene $s$ in the video $v$ we obtain a temporal ($\hat{t}$) and spatial ($\hat{x}$) feature embedding using 2D and 3D convolutional neural networks denoted here as $h(.)$ and $g(.)$. A shared positional embedding, $p$, is added to every scene embedding to generate the spatial tokens, $z_s$, and the temporal tokens, $z_t$. A two stream spatio-temporal transformer with two layers ($l$), learns the dependency between the sequence of spatial tokens and the temporal tokens. Following normalisation and a linear projection, the features are fused for classification. In practice, we only fuse and classify the prepended CLS token as discussed in Section \ref{method}. When training the STAN-Small model, we do not back-propagate through the 3D CNN represented here by the stop-gradient line. For STAN-Large we continue to fine-tune both convolutional encoders.}
   \label{fig:sampling}
 \end{figure}

\textbf{Overview}
We aim to classify long videos by splitting them into discrete scenes and extracting spatial and temporal representations for temporal fusion via a two-stream transformer encoder. To do so, we first analyse the video for changes in average frame intensity/brightness using a running average over RGB video channels. We use these time stamps as scene segmentation points and uniformly sample 12 frames from each scene segment using a 3D CNN encoder to generate a low-resolution temporal feature tokens for each scene. We also extract the central frame of each scene at a higher resolution and use a 2D CNN to obtain a spatial feature token. The spatial and temporal scene tokens are encoded with a positional embedding and temporally aggregated using a two-stream transformer encoder. For classification, we randomly initialise an additional token prepended to the spatial and temporal sequence of embedding tokens, which learns to model the temporal inter-dependency of the individual scenes. An overview of the sampling methodology is shown in Fig \ref{fig:sampling}.

\textbf{Spatial and Temporal Tokens} Given a uniformly sampled set of 12 video frames from a scene defined here as $t \in \mathbb{R}^{12 \times H \times W \times C}$ we implement a R(2+1)d Video ResNet encoder \cite{r21d} pretrained on the Kinetics400 Dataset \cite{carreira2017quo_kinetics}, defined as $g(t)$, to encode each set of frames into a single feature embedding token $\hat{t}$ which represents the projected temporal features of the scene $s_i$ from a set of scenes $s$. We obtain an embedding for each scene in the video using the above method to generate a set ${z_t}$ which represents the set of all temporal embedding vectors for the scene $s_i \in s$. 

\begin{equation}
z_t = [z_{cls}, \hat{t_0}, \hat{t_1}, ..., \hat{t_n}] + p
\label{eq:temporal_list}
\end{equation}

Where ${z}_{cls}$ is a randomly initialized vector in the same dimension as $\hat{t_i}$ and $p_i \in \mathcal{PE}$ is a positional embedding added to the sequence as described in \cite{attention_all_you_need} as,

\begin{equation}
\mathcal
{PE}_{pos,2_i} = sin(\frac{pos}{10000^{2i/d_{model}}})   \nonumber
\end{equation}

\begin{equation}
\mathcal
{PE}_{pos,2_{i+1}} = cos(\frac{pos}{10000^{2i/d_{model}}})
\end{equation}

Where $d_{model}$ is a common dimension for both the spatial and temporal tokens. In \cite{video-action-transformer,vidtr,16x16}, positional information is used to infer the relationship between cropped regions in an individual frame. As we have introduced inductive bias via convolution, we do not need to model the position of pixel regions and instead extend the positional embedding method to infer the position of each short temporal sequence. We argue that this is logical for longer videos that feature a narrative composed of multiple dynamic actions and environments such as those found in movies. In section \ref{sec:results} we show that this positional information benefits the model's performance on such a task.

\textbf{Spatial Encoding Token} To obtain a spatial encoding token, we perform a linear projection of the high resolution frame $x$, which is sampled from the center of $s_i$, to a spatial embedding token $\hat{x}$ using a ResNet18\cite{resnet} encoder model pretrained on the ImageNet Dataset \cite{imagenet_cvpr09} so that $\hat{x} = h(x)$.  As in the case of the temporal token, we form a sequence of spatial embedding tokens $z_x$ from spatial feature embeddings obtained via $h(.)$ so that $z_x = {\{\hat{x_0}, \hat{x_1}, ..., \hat{x_n}\}}$ where $n$ is the total number of scenes in the video $z$. In Eq \ref{eq:spatial-list} $z_x^{cls}$ is the randomly initialized spatial cls token used for classification. 

\begin{equation}
{z_s} = [z_x^{cls}, \hat{x_{0}}, \hat{x_{1}}, ..., \hat{x_{n}}] + p
\label{eq:spatial-list}
\end{equation}

The positional embedding vector $p$ in Eq \ref{eq:spatial-list} is the same as defined in Eq \ref{eq:temporal_list} and as such the positional information is shared between the spatial and temporal tokens. This ensures that positional information is consistent between the spatial and temporal streams during fusion. We now define a spatial and temporal transformer for temporal aggregation of the input tokens and then discuss techniques to fuse the output classification from both streams. 

\textbf{Spatial and Temporal Transformer Encoders} For the spatial and temporal transformer encoder architecture, we implement the transformer model introduced in \cite{attention_all_you_need} originally designed for natural language processing. As described in \cite{attention_all_you_need} we generate Query ($Q$), Key ($K$), and Value ($V$) matrices from both the spatial and temporal embedding representations where each row in the matrix represents a corresponding scene, with the first row representing our classification token. A matrix of outputs is computed as,

\begin{equation}
Attention(Q, K, V) = softmax(\frac{QK^T}{\sqrt{d_k}V})
\label{eq:attention}
\end{equation}

The output matrix $A(Q,K,V)$ is summed with the input embedding matrix via a residual connection and normalised using Layer Normalisation. Finally, the output is summed with a second residual connection before a classification MLP head. In practice, we only apply the final classification MLP to the temporal and spatial transformer classification tokens which learn a representation of the input via self-attention. As such, we can discard the other rows of the output matrix and only backpropagate via the classification MLP. We use Multi-Headed-Self-Attention to model further representations in spatial and temporal domains by replicating the attention mechanism in Eq \ref{eq:attention} and concatenating the heads. This process is described in detail in \cite{attention_all_you_need}. In \cite{video-action-transformer,vidtr,convit} the authors use a linear layer for classification based on ${z}_{cls} \in \mathbb{R}^d$. In our work, we linearly project ${z}_{cls}$ to $\mathbb{R}^{d}_{model}$ for both the spatial and temporal features so we can experiment with several fusion methods, which we will describe next. 

\textbf{Fusion and Classification} For fusion, we normalise the output embeddings and project to a common dimension with a linear layer before classification via a three-layer MLP separated by a gated non-linearity. The loss function can be defined as a binary cross entropy loss between the targets $y$ and the scaled sum of the feature embeddings,

\begin{equation}
   \mathcal {L}^{Fusion} = {L}_{BCE}(h(Norm([q(z_s^{cls}) + \lambda q(z_t^{cls})]), y)
\end{equation}

Where $\lambda < 1.0$ acts as a hyper-parameter to scale the influence of the temporal network in generating the output logits and set to 0.6. $q(.)$ Is an MLP with one hidden layer and weights shared for both the temporal and spatial streams, while the function $h(.)$ represents the final MLP used for classification after fusion. To improve training time we use transfer learning and pre-train the 2D CNN on ImageNet \cite{imagenet_cvpr09} and the 3D CNN on Kinetics400 \cite{carreira2017quo_kinetics}. We present both a large, and small version of the model. For the \textbf{STAN-Small}, we do not update the parameters in the temporal stream layers but back-propagate through the two-stream transformer to the 2D CNN. This reduces the number of trainable parameters by 48 Million, making the whole network trainable on a single Nvidia RTX5000 GPU with 16GB of memory. The larger model, \textbf{STAN-Large}, which back-propagates via both streams, improves performance by 11\% but requires 92.5 million trainable parameters compared with 45 million for STAN-Small. In Table \ref{tab:main-results} we experiment with a number of additional methods for fusion including collaborative gating \cite{liu2019CollabGate} and distillation \cite{hinton2015distilling}. A full experimental analysis follows in the next section.

\section{Results}
\label{sec:results}
We evaluate the proposed method on several long video classification tasks. As discussed in \cite{wu2021towards} video datasets have typically focused on short video tasks, therefore following~\cite{wu2021towards} we define long videos as videos which feature more than three shots or scenes, are longer than a minute in length, and in which classification relies on, or is improved by, contextual understanding of the relationship between the content of shots and their order. To evaluate our method, we use both the \textbf{MMX-Trailer-20 Dataset (MMX)} \cite{rethinking-genre}, and the \textbf{Long Video Understanding Benchmark (LVU)}~\cite{wu2021towards}.

\textbf{MMX Dataset} The MMX-Trailer-20 Dataset \cite{rethinking-genre} is a multi-modal movie trailer dataset with 7555 movie trailers spanning many scenes, locations, actions, and narratives. The MMX-Trailer dataset features movie trailers from 20 genres with six genre labels per video from 1900 $-$ 2021 with an average length of two minutes. The training set is 6047 videos, with the validation and test sets both at 754 samples. The dataset consists of ~37 million frames in total. 

\textbf{LVU Benchmark} The Long Video Understanding Benchmark \cite{wu2021towards} contains ~30K videos with an average length of ~120 seconds. The benchmark comprises of content understanding, user engagement prediction, and movie metadata prediction tasks, demonstrating various requirements for long temporal modelling. We test our approach on the content understanding tasks including character relationship identification, speaking style, and scene recognition. 

\textbf{Compared Methods} To demonstrate the effectiveness of the two-stream network, we implement several existing methods for video classification using the MMX-Trailer-20 dataset. They include extracting convolutional image features from a ResNet18 \cite{resnet} for each scene and simply using average pooling for classification. We also implement this same method with features obtained from a two-stream Inflated 3D Convolutional Neural Network \cite{3dconv} and a SENet \cite{squeeze}. We also show results for comparative temporal networks, including a Temporal Pyramid Network, which extracts multiple level features from a CNN to model dynamic temporal movements in videos, and a vanilla LSTM \cite{lstm} to model the temporal relationship between the feature representations from an 18 layer ResNet. More complex temporal aggregation strategies are also explored, such as the Fine-Grained Semantic \cite{rethinking-genre} architecture which uses concatenation to aggregate multi-modal features as well as the effectiveness of earlier audio-visual works such as audio-visual classification using support vector machines \cite{huang2012movie} and audio VGG network as described in \cite{youtube-8m}. We also explore distillation as described in \cite{DEITT}, using the temporal transformer encoder as a teacher network to the spatial transformer. For the LVU Benchmark task we compare our method with those presented in the paper including videoBERT \cite{videobert}, R101-SlowFast \cite{slowfast} and Object Transformer \cite{wu2021towards}.

\textbf{Metrics} As outlined in \cite{rethinking-genre} the MMX dataset is imbalanced and as such we follow other works \cite{dong2016word2visualvec,Miech2017,mithun2018learning} and use Mean Average Precision $mAP$ to evaluate the effectiveness of our classifier. To calculate the $mAP$, we average the area under the precision-recall curve per genre, weighting instances according to the class frequencies. We also show weighted Precision ($P_w$), weighted Recall ($R_w$), and weighted F1-Score ($F1_w$) in Table \ref{tab:genre-specific-results}. With all metrics, a higher value demonstrates improved accuracy. For LVU we compute the standard error averaged over five runs as proposed in the original paper \cite{wu2021towards} where higher values represent improved performance.

\textbf{Evaluation} First, in Table \ref{tab:main-results} we evaluate our approach against existing methods for video classification using the MMX-Trailer-20 Dataset. We demonstrate that our method outperforms pooling of convolutional spatial features by 10\%. Secondly, we show our method improves performance on scene level features extracted via an inflated 3D CNN \cite{x3d} by 11\%. Third, we compare our approach with other methods for combining convolutional features temporally, including an LSTM \cite{lstm}, S(TPN) \cite{s-tpn}, concatenation \cite{rethinking-genre}, and collaborative gating \cite{liu2019CollabGate}. Finally, we show results for an alternative method for two-stream aggregation using a distillation network as described in \cite{DEITT}. We outperform all existing methods for the genre classification task.

\begin{table}

\centering
\begin{adjustbox}{width=1\textwidth}
\begin{tabular}{l|ccc|c}
\toprule
Method                       &       CNN Feature Extractor & Frames per Scene  & Aggregation Method & $\overline{mAP}$   \\
\midrule
ResNet \cite{resnet} &  ResNet18 & 1 & Avg Pool & 0.434 \\
SqueezeExcite~\cite{squeeze} & SE-ResNet & 1 & Avg Pool &0.544 \\
\midrule
I3D~\cite{carreira2017quo_kinetics} & I3D & 12 & Avg Pool & 0.487 \\
\midrule
Collaborative Gating \cite{liu2019CollabGate} & ResNet18 & 12 & Gated Unit & 0.4723 \\
S(TPN) \cite{s-tpn}  & ResNet18 & 12 & Temporal Pyramid &0.492          \\ 
Fine-Grained Semantic \cite{rethinking-genre}  & Multi-Modal & 16 &  Concatenation &   0.583 \\
LSTM \cite{lstm} &  ResNet18 & 1 & LSTM & 0.596 \\
Distillation \cite{hinton2015distilling,DEITT}  & ResNet18 + R2+1D & 12 & Transformer &  0.601 \\
\midrule
\textbf{STAN-Small}  & ResNet18 + R2+1D & 12 &  Transformer & \textbf{0.640}     \\
\textbf{STAN-Large}  & ResNet18 + R2+1D & 12 &  Transformer & \textbf{0.750}     \\
\bottomrule
\end{tabular}

\end{adjustbox}
\bigskip
\caption{Comparison of our proposed approach with existing methods for video classification using CNN feature extractors and evaluated on the MMX-Trailer-20 Dataset. We implement several feature extraction and aggregation methods to evaluate their effectiveness for the long video classification task.}
\label{tab:main-results}

\end{table}

\begin{table}

\centering
\begin{tabular}{@{}l|lll@{}}
\toprule
Method & \multicolumn{1}{c}{Relation} & \multicolumn{1}{c}{Speaking} & \multicolumn{1}{c}{Scene} \\
\toprule
R101-SlowFast+NL \cite{slowfast}   & 52.4  & 35.8  & 54.7  \\
VideoBERT \cite{videobert}         & 52.8  & 37.9  & 54.9  \\
Object Transformer \cite{wu2021towards} & 53.1  & 39.4  & 56.9  \\
STAN-Large (ours)     & \textbf{56.25} & \textbf{41.41} & \textbf{58.33} \\
\bottomrule

\end{tabular}
\bigskip
\caption{Accuracy of our approach on long video understanding tasks using the Long Video Understanding Dataset. The reader can find references and further details in \cite{wu2021towards}. We outperform current approaches for classifying conversation (speaking), character relationships (relation), and locations of scenes (scene).}
\label{tab:lvu}

\end{table}
In Table 2 we show that we achieve SOTA results for three long-form video understanding tasks on the LVU Benchmark outperforming other methods which also utilise transformer architectures such as VideoBERT \cite{videobert} and \cite{wu2021towards} which utilises pre-trained CNN backbones plus self-supervised masked pre-training. Our architecture performs particularly well on the relationship identification task $(\bf{+3.1})$ despite having no prior knowledge of the domain in pre-training. 

\textbf{Ablation Experiments} In Table~\ref{tab:ablation} we perform several ablation studies to assess the impact of the spatial and temporal features and the data efficiency of the model. To measure the data efficiency, we randomly sampled 2000 trailers from the MMX-Trailer-20 training partition, with the entire test partition of 754. Table~\ref{tab:ablation} shows that the model still achieves SOTA performance on the MMX-Trailer-Dataset despite being trained on only a quarter of the samples demonstrating high data efficiency. We infer that data efficiency is improved by using convolutional encoding as the transformer network only needs to map self attention between the scene feature tokens rather than pixel localities. Furthermore, the proposed network architecture can infer translation in-variance within the convolutional encoding.

In Table \ref{tab:ablation} we also provide further results for models that only use either the spatial or temporal convolutional encoder to assess the impact of propagating gradients through the convolutional encoders.  We observe that back-propagating through the spatial convolutional encoder provides the most significant performance gain with the most negligible effect on the number of trainable parameters. Training both the CNN encoders end to end (STAN-Large) is the most effective method for achieving high accuracy. Still, it comes at a cost, increasing the number of trainable parameters by 48 million. We find that using just the temporal transformer encoder and introducing a stop gradient before the convolutional feature extractor performs well but is improved with spatial features for 28 million additional trainable parameters. We also show genre specific results in Table \ref{tab:genre-specific-results}.

\begin{table*}[htbp]

\centering
\begin{adjustbox}{width=1\textwidth}
\begin{tabular}{l|cccccccccccccccccccc|cccc}
Model                                 &Actn &Advnt&Animtn&Bio &Cmdy&Crme&Doc &Drma&Famly&Fntsy&Hstry&Hrror&Mystry&Music&SciFi&Sprt &Shrt&Thrll&War&$F1_w$  &$mAP$&$P_w$     & $R_w$\\ 
\midrule
Support                               & 130 &197  &46    &13  &224 &102 &87  &267 &117  &115  &44   & 104 &41   &86   &107   &30   &45  &12   &21  & -     & -                & -        & -\\
\midrule
Random                                &0.29 &0.41 &0.11  &0.03&0.46&0.24&0.21&0.52&0.27 &0.26 &0.11 & 0.24&0.1   &0.2  &0.25  &0.08 &0.11&0.03 &0.05&0.318  &    	0.134	  &0.19	     &1   \\
ResNet~\cite{resnet}               &0.43&0.55  &0.74  &0   &0.49&0.38&0.63&0.55&0.51 &0.28 &0.24 &0.42 &0.3	 &0.28&0.41  &0.22 &0.19&0.11 &0.33&0.434  &	0.489	         &0.437     &	0.48    \\        
VGG-Audio~\cite{youtube-8m}         &0.47&0.51  &0.40  &0.10&0.61&0.38&0.58&0.55&0.51&0.37 &0.11  &0.34 &0.39  &0.30&0.35   &0.16 &0.15&0.13 &0.12&0.454  &	0.449	         &  0.400   &0.537    \\
I3D~\cite{carreira2017quo_kinetics}&0.5 &0.59  &0.74	 &0   &0.62&0.33&0.63&0.56&0.55 &0.36 &0.2  &0.38&0.45  &0.24&0.37  &0.23 &0.14&0.10 &0.13&0.463  &	0.487	         &0.448	    &0.494   \\
SqueezeExcite~\cite{squeeze}            &0.48&0.63  &0.79  &0.12&0.65&0.41&0.60&0.59&0.55&0.42 &0.25  &0.47&0.42  &0.29&0.50  &0.34 &0.19&0.12 &0.31&0.516  &	0.554            &0.493     &0.572    \\
Naive Concat \cite{resnet}                             &0.56&0.61  &0.64  &0.09&0.64&0.35&0.69&0.60&0.58&0.39&0.19&0.49 &0.45 &0.21    &0.48 &0.39 &\textbf{0.28} &0.27&	0.41&0.525 &0.497             &0.522    &0.551\\
Fine-Grained Semantic \cite{rethinking-genre} &{0.62}&\textbf{0.69}&{0.71}&0.11&\textbf{0.71}&{0.53}&\textbf{0.73}&{0.62}&{0.51}&{0.34}&{0.56}     &\textbf{0.60}&\textbf{0.45} &0.50&\textbf{0.64}&\textbf{0.30}&0.11&{0.13}&\textbf{0.55}&{0.597}&{0.583}&{0.554}&{0.697} \\ 
\midrule
\textbf{STAN-Small}  &\textbf{0.71}&{0.68}&\textbf{0.92}&\textbf{0.21}&{0.61}&\textbf{0.65}&{0.62}&\textbf{0.69}&\textbf{0.86}&{0.49}&\textbf{0.46}&\textbf{0.58}&{0.43}&{0.39}&\textbf{0.53}&{0.13}&{0.20}&\textbf{0.85}&{0.50} & \textbf{0.65} & \textbf{0.64} & \textbf{0.62} & \textbf{0.73} \\ 
\bottomrule
\end{tabular}
\end{adjustbox}
\bigskip
\caption{
Genre classification performance for each genre on the MMX-Trailer-20 dataset. We observe high performance gains on genres in which temporal information can be considered an important classifier such as Action $+9$ and Animation $+11$. We also observe that other network architectures perform very poorly on classification of the genre thriller while we improve accuracy by $+58$. We conclude that long-term temporal modelling performs well on this task as the content is difficult to classify when features are presented in isolation.}
\label{tab:genre-specific-results}

\end{table*}

\begin{table}[]
\centering

\begin{tabular}{l|ccc}
\toprule
Method                           & Samples & Parameters (Millions) &mAP\\
\midrule
Spatial with backprop     & 2000 &28 & 0.5903        \\
Temporal with backprop     &     2000  &48 &  0.6221         \\
\midrule
Spatial  no backprop            &2000  &16.5 & 0.4024 \\
Temporal no backprop        &2000  &16.5 & 0.59 \\
\midrule
Distillation Network         &2000 &44.5 &  0.6005           \\
Gated Fusion            &2000    &45 & 0.4728             \\
\midrule
STAN-Small          &2000 &45 &   0.6151          \\
STAN-Small       &6047 &45 &  0.6401 \\     
STAN-Large       &6047 &92.5 &  0.7506 \\   
\bottomrule
\end{tabular}
\bigskip
\caption{Ablation experiments assess the network quality under a constrained data training protocol. Each model is trained using only 2000 samples, while the test length remained consistent at 754 samples. The network continues to outperform existing methods with fewer data. We also show results for individual spatial and temporal streams with back-propagation. }
\label{tab:ablation}
\end{table}

 \begin{figure}[!h]
  \includegraphics[width=1\textwidth]{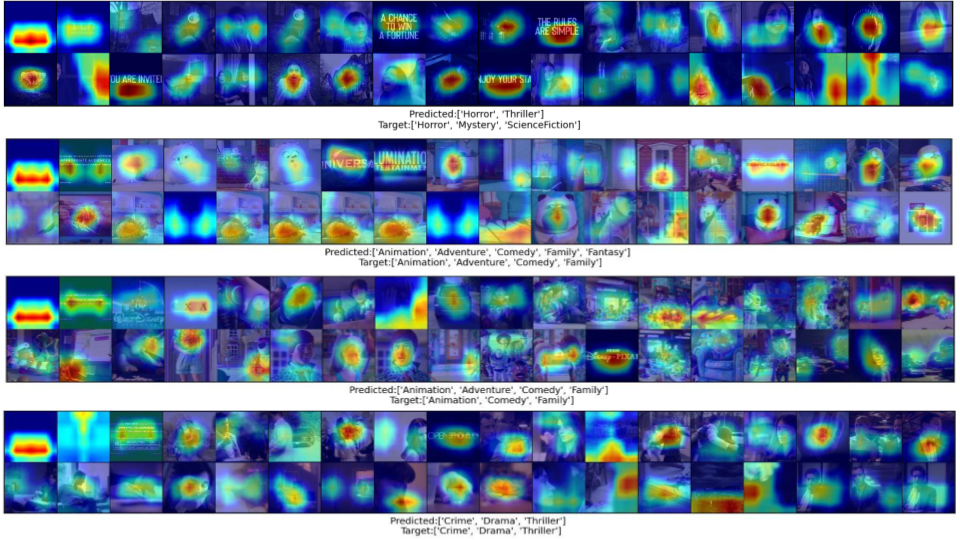} 

  \caption{Class activation maps of randomly selected samples from the MMX-Trailer-20 test partition. Each sub-figure represents a series of input tokens from the spatial encoder $h$. The class activation maps show pixel regions in red to blue, with red regions representing high activation and, therefore, greater contribution to the predicted class labels.}
  \label{fig:activation}
 \end{figure}

 

\textbf{Qualitative Results}  In Fig \ref{fig:activation} we show class activation maps of the input 2D CNN encoder, shown in Fig \ref{fig:sampling} as $h()$ and the predicted genres for a given input video. We observe that the model predicts the labels correctly and learns cohesive features for the input tokens. For example in Fig \ref{fig:activation} we see that the Family genre is predicted when we have a strong activation on animals faces. For the Horror classification, spatial features include people screaming and dark scenes. We also notice that text is an important feature for the classification task, acting as a strong temporal marker in the sequence of tokens. Class activation maps are obtained via Grad-CAM\cite{gradcam-paper} using the code provided by \cite{gradcam}. We refer the reader to the appendix for further qualitative results and ablation experiments.
 
\section{Conclusions}

We present a data and memory efficient spatio-temporal attention network for long video classification which combines the advantages of convolutional inductive bias with the computational advantages of transformer networks for the task of long video classification. We show that by using static \textit{image} and temporal \textit{context} convolutional tokens we are able to create a data efficient architecture capable of classifying videos of up to two minutes in length and trained on a single GPU while achieving SOTA results on a range of long-form video understanding tasks. 
\bibliography{egbib.bib}
\end{document}